\pdfoutput=1
\documentclass[11pt]{article}

\usepackage{acl}
\usepackage{times}
\usepackage{latexsym}
\usepackage[T1]{fontenc}
\usepackage[utf8]{inputenc}
\usepackage{microtype}
\usepackage{graphicx}
\usepackage{subfigure}
\usepackage{tipa}
\usepackage{amsmath}
\usepackage{amssymb}
\usepackage{xcolor}

\title{Language-Agnostic Meta-Learning for Low-Resource \\Text-to-Speech with Articulatory Features}

 \author{Florian Lux \and Ngoc Thang Vu \\
         Institute for Natural Language Processing \\ University of Stuttgart\\ \texttt{florian.lux@ims.uni-stuttgart.de}}

\begin{document}
\maketitle
\begin{abstract}
  While neural text-to-speech systems perform remarkably well in high-resource scenarios, they cannot be applied to the majority of the over 6,000 spoken languages in the world due to a lack of appropriate training data. In this work, we use embeddings derived from articulatory vectors rather than embeddings derived from phoneme identities to learn phoneme representations that hold across languages. In conjunction with language agnostic meta learning, this enables us to fine-tune a high-quality text-to-speech model on just 30 minutes of data in a previously unseen language spoken by a previously unseen speaker.
\end{abstract}

\section{Introduction}
The advance of deep learning \cite{vaswani2017attention, goodfellow2014generative} has enabled great improvements in the field of Text-to-Speech (TTS). (Towards-)end-to-end models, such as Tacotron 2 \cite{wang2017tacotron, Shen/etal:2018}, TransformerTTS \cite{li2019neural}, FastSpeech 2 \cite{ren2019fastspeech, ren2020fastspeech}, FastPitch \cite{lancucki2021fastpitch} and many more famous instances (e.g. \citet{Arik/etal:2017} and \citet{Prenger/Valle/Catanzaro:2019}) allow for speech synthesis with unprecedented quality and controllability. The models mentioned here rely on vocoders, such as WaveNet \cite{VanDenOord/etal:2016}, MelGAN \cite{kumar2019melgan}, Parallel WaveGAN \cite{Yamamoto/Song/Kim:2020} or HiFi-GAN \cite{kong2020hifi} to turn the parametric representations that they produce into waveforms. Recently proposed models even include some with the ability to go directly to the waveform from a grapheme or phoneme input sequence, such as EATS \cite{donahue2020end} or VITS \cite{kim2021conditional}. 

While these methods all perform remarkably well if given enough data, cross-lingual use of data remains a key challenge in TTS. Most modern methods are limited to languages and domains that are rich in resources, which over 6,000 languages are not. Attempts at reducing the required resources in a target language by making use of transfer learning from multilingual data have been made by \citet{9208651, 3403331, tu2019endtoend}. The mismatch of input spaces however requires complex architectural changes, which limits their ability to be used in conjunction with other modern TTS architectures. Attempts at fixing the issue of having to transfer knowledge from a source to a target by just jointly training on a mixed set of more and less resource rich languages have been made by \citet{he2021multilingual, dekorte2020efficient, yang2020towards}, which requires complex training procedures. In this work, we will also attempt to transfer knowledge from a set of high resource languages to a low resource language. We fix previous shortcomings by 1) using a linguistically motivated representation of the inputs to such a system (articulatory and phonological features of phonemes) that enables cross-lingual knowledge sharing and 2) applying the model agnostic meta learning (MAML) framework \cite{finn2017modelagnostic} to the field of low-resource TTS for the first time.

Using articulatory features as inputs for neural TTS has been attempted recently by \citet{Staib_2020} and \citet{dan}, following the classical approach of \citet{jakobson1951preliminaries}. Both achieved good results when applying this idea to the codeswitching problem, since unseen phonemes in the input space no longer map to nonsensical positions, as it would be the case for the standard embedding-lookup. It has to be noted however, that this only works across languages with similar types of phonemes. Also \citet{gutkin2017uniform} have applied phonological features to low-resource TTS with fair success. They did however rely on supplementary features, such as dependency parsers and morphological analyzers. Furthermore all of their data and models are proprietary and can therefore not be used to compare results to. In this work, we extend the use of articulatory inputs with the MAML framework to enable very simple yet well working low-resource TTS that can be applied to almost all modern TTS architectures.

We encounter severe instabilities when using MAML on TTS, which make the standard formulation of MAML infeasible to use. Thus we also propose a modification to MAML, which reduces the procedure's complexity. This allows us to create a set of parameters of a model that can be used to fine-tune to a well working single-language single-speaker TTS model with as little as 30 minutes of paired training data available and even enables zero-shot adaptation to unseen languages. We evaluate the success of our approach with both automatic measures and human evaluation. 

Our contributions are as follows: 1) We show that it is beneficial to train a TTS model on articulatory features rather than on phoneme-identities, even in the standard single-language high-resource case; 2) We introduce a training procedure that is closely related to MAML which allows training a set of parameters for a TTS model that can be fine-tuned in a low resource scenario; 3) We provide insights on how much data and training time are required to fine-tune a model across different languages and speakers simultaneously using said meta-parameters; 4) We show that the meta-parameters can generalize to unseen phonemes and rapidly improve their ability to properly pronounce them when fine-tuning.
\footnote{All of our code, as well as the checkpoints for a low-resource fine-tuning capable Tacotron 2 and FastSpeech 2 model are publicly available at \url{https://github.com/DigitalPhonetics/IMS-Toucan}.}

\section{Background and Related Work}
\subsection{Input Representations}
\paragraph*{Character Embeddings} The simplest approach to representing text as input to a TTS is using indexes of graphemes to look up embeddings. This is however prone to mistakes. \citet{taylor2020enhancing} bring up the example of \textit{coathanger}. If the TTS is not aware of the morpheme boundary between the \textit{coat} and the \textit{hang}, it will be inclined to produce something like \mbox{[k\textturnv \texttheta \textschwa \textsci n\textdyoghlig \textschwa]} rather than the correct \mbox{[ko\textupsilon th\ae \textipa{N}\textschwa]}. Such a representation of the input will be highly language dependent, since special pronunciation rules rarely hold for more than a single language. 

The textual input can be augmented by adding information, such as morpheme boundaries, intonation phrase boundaries derived from e.g. syntactic parsing as is done in many TTS frontends \cite{schroder2003german, clark2007multisyn, ebden2015kestrel}, or even the semantic identity of the word a character belongs to, using e.g. BERT embeddings \cite{hayashi2019pre}. 

\paragraph*{Phoneme Embeddings} Rather than looking up embeddings for graphemes, it is often beneficial to use embeddings of phonemes. Phonemizers \cite{bisani2008joint, taylor2005hidden, rao2015grapheme} produce a sequence of phonetic units, which correlate with the segments in the audio much more than raw text. One such standard of phonetic representation which we make use of is the International Phonetic Alphabet (IPA). Using this set of phonetic units alleviates the problems of TTS fine-tuning and transfer-learning to low-resource domains, because the phonetic units should be mostly language independent. \citet{deri2016grapheme} provide a data driven approach for the grapheme to phoneme conversion task, which performs well on over 500 languages and can be adapted fairly easily to any new low-resource language. There remains however one major challenge: The use of different phoneme sets for each language, leading to completely unseen units in inference or fine-tuning data.

\paragraph*{Latent Representations} \citet{li2019bytes} claim that multilinguality in speech recognition and TTS can be achieved by changing the input to a latent representation that is trained across languages. While their results seem very promising, their technique needs training data in all languages it should be applied to, which rules out zero-shot settings.

\paragraph*{Articulatory Features} We fix the shortcoming of not being able to handle unseen phonemes by specifying phonemes in terms of articulatory features such as position (e.g. frontness of the tongue) and category (e.g. voicedness). We show that systems trained on this input can produce a phoneme given nothing but an articulatory description and thus generalize to unseen phonemes. This makes the transfer of knowledge across languages much simpler. A similar approach for the purpose of handling codeswitching has been done in \citet{Staib_2020}. Our work builds on top of theirs by extending the idea to transfer learning an entire TTS in a new language with minimal data, making use of meta learning on top of articulatory features.

\subsection{Model Agnostic Meta Learning (MAML)} 
The goal of MAML \cite{finn2017modelagnostic} is to find a set of parameters, that work well as initialization point for multiple tasks, including unseen ones. The procedure consists of an outer loop and an inner loop. The outer loop starts with a set of parameters, which we will call the Meta Model. The inner loop trains task specific copies of the Meta Model for a low amount of steps. Once the inner loop is complete, the loss for each of the models is calculated, summed, and backpropagated to the original Meta Model by unrolling the inner loop. This includes the very costly calculation of second order derivatives. The Meta Model is then updated and the inner loop starts again.

This procedure moves the initialization point closer to the optimal configuration for each of the trained tasks, which generalizes to even unseen tasks. Multiple variants of MAML have been suggested that try to fix the high computational cost of the second order derivatives. The simplest one is called first-order MAML and simply applies the gradient of the task specific model at the end of the inner loop directly to the Meta Model. Other variants are described in \citet{antoniou2018train, rajeswaran2019meta}.

\section{Approach}
\subsection{System Description}
\begin{figure}[t]
    \centering
        \includegraphics[width=\linewidth]{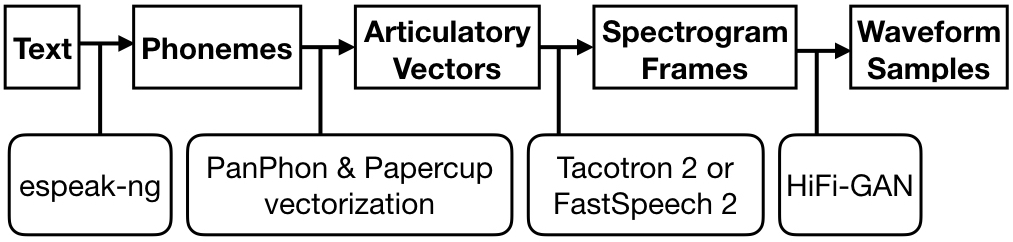}
    \caption{Overview of the TTS pipeline we use. The top row shows the modality in which the data is at this point in the pipeline. The lower row shows the methods that handle the transitions. Each of the blocks in the lower row can be exchanged easily with other methods that have the same interfaces.}
    \label{sys}
\end{figure}

For the implementation of our method, we use the open source IMS Toucan speech synthesis toolkit, first introduced in \cite{Lux2021TheIT}, which is in turn based on the ESPnet end-to-end speech processing toolkit \cite{watanabe2018espnet, hayashi2020espnet, hayashi2021espnet2}. \citet{neekhara2021adapting} show, that it is beneficial to fine-tune a single-speaker model to a new speaker rather than to train a multi-speaker model. Inspired by this, we decided to also use a model that is not conditioned on speakers or on languages rather than a conditioned multi-speaker multi-lingual model and fine-tune it on the data from a new speaker in a new language. In preliminary experimentation we got similar results to them within one language, but found their method to not work across languages. In comparison to the fine-tuning of a simple single speaker model, we found training and fine-tuning a model conditioned on language embeddings and speaker embeddings much more sensitive to the choice of hyperparameters. Figure \ref{sys} shows an overview of our system, underlining how it is not specific to a certain architecture, but could instead be used in conjunction with almost all modern TTS methods.

\paragraph*{Tacotron 2} For our implementation of Tacotron 2 \cite{Shen/etal:2018}, we make use of the forward attention with transition agent introduced in \citet{zhang2018forward}, which uses a CTC-like forward variable \cite{graves2006connectionist} to promote the quick learning of monotonic alignment between text and speech. To further help with this, we make use of the guided attention loss introduced in \citet{tachibana2018efficiently}.

\paragraph*{FastSpeech 2} To train the parallel FastSpeech 2 model \cite{ren2020fastspeech}, annotations of durations for each phoneme are needed. These also have to be generated for the low-resource fine-tuning data. To that end, we generate alignments using the encoder-decoder attention map of a Tacotron 2 model. Following \citet{kim2020glow, shih2021rad, badlani2021one}, we apply the Viterbi algorithm to find the most probable monotonic path through the attention map, which significantly improves the quality of the alignments.  

This is especially important, because we train our FastSpeech 2 model with pitch and energy labels that are averaged over the duration of each individual phoneme to allow for great controllability during inference, as is introduced by \citet{lancucki2021fastpitch}. Incorrect alignments would lead to follow-up errors such as an unnaturally flat prosody. 

Furthermore, we make use of the conformer block \cite{gulati2020conformer} as the encoder and decoder, rather than the standard transformer \cite{vaswani2017attention}.

\subsection{Articulatory Vectors}
\paragraph*{PanPhon} The PanPhon resource \cite{Mortensen-et-al:2016} can be used to get linguistic specifications of phonemes. It comes with an open-source tool\footnote{\url{https://github.com/dmort27/panphon}} which we use to convert phonemes into numeric vectors. Each vector encodes one feature per dimension and takes the value of either -1, 0 or 1, putting the features on a scale wherever meaningful. This featureset also includes phonological features which go beyond simple phonetics, such as whether a phoneme is syllabic.

\paragraph*{Papercup} Additionally we make use of the purely articulatory description system of phonemes introduced in \citet{Staib_2020}, which we will call Papercup features in the following. 
For the encoding we use one-hot vectors, similar to their implementation. Some of the features, like openness or frontness, should be on a scale rather than one-hot encoded. However since the articulatory vector is fed into a fully connected layer, we leave the reconstruction of this dependency between features for the network to learn.

\subsection{Language Agnostic Meta Learning}
We find that the standard implementation of MAML does not work well for the TTS task. The inner loop needs hundreds of updates in order to make a significant change to the performance of the task specific model. This is probably due to the TTS task being a one-to-many mapping task, where the loss function of measuring the distance to a spectrogram is not an accurate objective for the TTS. For every text, there are infinitely many spectrograms, which could be considered gold data. Those spectrograms could differ in e.g. the speaker who reads the text and how they read the text. Since there are no conditioning signals, the TTS has to update its parameters towards a certain speaker's characteristics in general. However because in our case each task is a different language and a different speaker, the training becomes highly unstable. So ideally we would either need to run MAML's inner loop until convergence, which is generally infeasible, or stabilize the procedure by not allowing the model to adapt further to one task than to the others.

To fix this issue, we calculate the Meta Model's loss on one batch per language. We then sum up the losses, backpropagate and update the Meta Model directly using Adam \cite{kingma2017adam}. This stabilizes the learning procedure, but still allows the model to update its parameters towards a more universal configuration. 
Since we have to make this simplification to MAML in order to deal with the different languages as tasks, we call this procedure language agnostic meta learning (LAML). Ultimately, the model should not care about the language it is fine-tuned in, since it should be close to a universal representation of an acoustic model. To give an exact notion of our modifications: We simplified equation \ref{eq:maml} to equation \ref{eq:laml}, where $opt$ is a gradient descent update, $B_i$ is a batch sampled from task $i$, $\mathcal{L}$ is an objective function, $\Theta$ is the set of parameters from the Meta Model and $\theta_i$ is the set of parameters specific to task $i$. To the best of our knowledge, we are the first to successfully apply MAML to TTS with languages being the tasks.
\begin{equation}
\begin{split}
    &\text{for $t$ steps do:} \\
    & \hspace{.4cm} \Theta_t = opt\left(\Theta_{t-1}, \nabla \sum_i \mathcal{L}\left(\theta_{i,d}, B_i\right)\right)\\
    &\text{where }\theta_{i,d=0} = \Theta_{t-1} \text{ and for $d$ steps do:}\\
    & \hspace{.4cm} \theta_{i, d} = opt \left(\theta_{i,d-1}, \nabla \mathcal{L}\left(\theta_{i, d-1}, B_i\right)\right)
    \label{eq:maml}
\end{split}
\end{equation}
\begin{equation}
\begin{split}
    &\text{for $t$ steps do:} \\
    & \hspace{.4cm} \Theta_t = opt\left(\Theta_{t-1}, \nabla \sum_i \mathcal{L}\left(\Theta_{t-1}, B_i\right)\right)
\end{split}
    \label{eq:laml}
\end{equation}

\section{Experiments}
In this section we will go over the experiments we conducted. First we will evaluate the articulatory features on their own in a single language setting using automatic measures. Then we will evaluate the combination of LAML and articulatory features in a cross-lingual setting using both automatic measures and human evaluation.

In our experiments we make use of the following datasets: The English Nancy Krebs dataset (16h) from the Blizzard challenge 2011 \cite{wilhelms2011lessac, king2011blizzard}; The German dataset of the speaker Karlsson (29h) from the HUI-Audio-Corpus-German \cite{puchtler2021huiaudiocorpusgerman}; The Greek (4h), Spanish (24h), Finnish (11h), Russian (21h), Hungarian (10h), Dutch (14h) and French (19h) subsets of the CSS10 dataset \cite{css10}.

\subsection{Mono-Lingual Experiments}
\subsubsection{Embedding Function Design} 
To explore our first hypothesis, we investigate the capabilities of the articulatory phoneme representations to be used in a single-speaker and single-language TTS system. To compare different ways of embedding the features, we train only the embedding function. As gold data we use the embeddings from a well trained lookup-table based Tacotron 2 model. In table \ref{tab:dist} we show the average distances of all articulatory vectors as projected by the embedding function to their identity based embedding counterpart. The distance $d$ between two embedding vectors $A$ and $B$ is defined in equation \ref{eq:distance}. 
\begin{equation}
    d = \left( \sum_i | A_i - B_i | \right) - \cfrac{\sum_i A_i \cdot B_i}{\sqrt{\sum_i A_i^2} \cdot \sqrt{\sum_i B_i^2}}
    \label{eq:distance}
\end{equation}
This distance function is also used as the objective function. The embedding functions are each trained for 3000 epochs using Adam \cite{kingma2017adam} with a batchsize of 32. The first column shows the results of the articulatory features being fed into a linear layer that projects them into a 512 dimensional space. The second column shows the results of the articulatory features being fed into a linear layer that projects them into a 100 dimensional space, applies the tanh activation function and then further projects them into a 512 dimensional space. 
As can be seen from the results, it is beneficial to both concatenate the PanPhon features with the Papercup features despite their overlap and to add a nonlinearity into the embedding function to match the embeddingspace of a well trained Tacotron 2 model. Hence we use this setup in all following experiments. 
\begin{table}[th]
    \centering
        \begin{tabular}{l|l l}
            $d$ & Linear & Non-Linear \\
            \hline
            PanPhon       & 0.47 & 0.1 \\
            Papercup      & 0.44 & 0.05 \\
            Combined      & 0.4  & \textbf{0.001}
        \end{tabular}
    \caption{Average distance of all embedded articulatory vectors to their position in an embedding space learned in a lookup-table based model.}
    \label{tab:dist}
\end{table}

\subsubsection{Convergence Time} 
\begin{figure}[th]
    \centering
    \subfigure[Proposed Tacotron 2 with articulatory features at 500 steps with a batchsize of 32.]{\includegraphics[width=.9\linewidth]{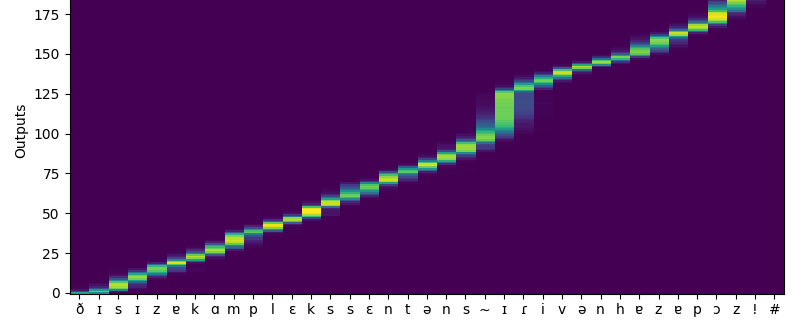}} \\
    \subfigure[Baseline Tacotron 2 with embedding-lookup at 2000 steps with a batchsize of 32.]{\includegraphics[width=.9\linewidth]{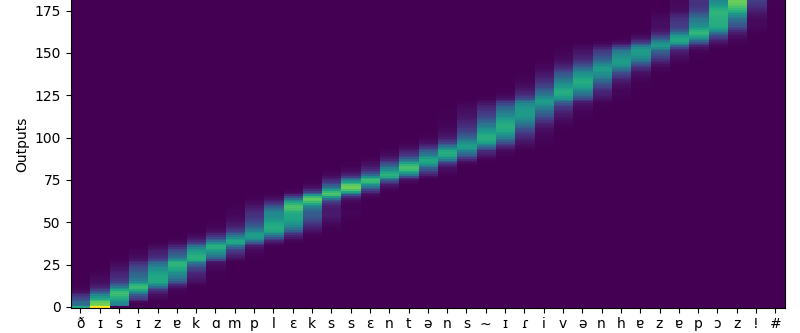}}
    \caption{The first instance of diagonal encoder-decoder attention on an unseen test sentence.}
    \label{fig:train}
\end{figure}

To investigate the impact that the articulatory features have on their own, we train a Tacotron 2 with and without them on the Nancy dataset and compare their training time and final quality. While the model trained on embedding tables shows a clear diagonal alignment of text and spectrogram frames on an unseen test sentence after 2,000 steps, the one trained on articulatory features does so already at 500 steps. This is visualized in figure \ref{fig:train}. The decoder of the Tacotron 2 model can only start to learn to decode after the alignment of inputs to outputs is learned.
So learning the alignment earlier gives the articulatory model a clear benefit. After training for 80,000 steps however, our own subjective assessment finds no difference in quality between the two. The earlier convergence of the alignment however shows a possible advantage of using the articulatory features on low-resource tasks, as quicker training progress means that training can be stopped earlier, before overfitting on little data becomes too problematic.

\subsection{Cross-Lingual Experiments} 
In order to investigate the effectiveness of our proposed LAML procedure, we train a Tacotron 2 model and a FastSpeech 2 model on the full Karlsson dataset as a strong baseline. We also train another Tacotron 2 model and another FastSpeech 2 model on speech in 8 languages with one speaker per language (Nancy dataset and CSS10 dataset) and fine-tune those models on a randomly chosen 30 minute subset from the Karlsson dataset. To our surprise, we did not only match, but even outperform the model trained on 29 hours with the model fine-tuned on just 30 minutes in multiple metrics. 

As a second baseline we tried to train another meta-checkpoint using the embedding lookup-table approach to also further investigate the effectiveness of the articulatory features. We did however not manage to get such a model to converge to a usable state. This already shows the superiority of the articulatory feature representations for such a multilingual use-case.

Furthermore we tried to fine-tune the well trained English single speaker models from the first experiment on the 30 minutes of German to have another baseline that can be used to measure the impact of the LAML procedure. This setup however also did not yield any usable results. During the fine-tuning process, the model was capable of speaking German with a strong English accent, yet it did not properly learn to speak in the voice of the target speaker. By the time the model learned to speak in the new speaker's voice, it had overfitted the 30 minutes of training data and collapsed, producing no more intelligible speech. We conclude that the method proposed in this paper not only improves on the ability to use cross-lingual data easily, but actually enables it in the first place. Both the articulatory features, as well as the LAML pretraining seem necessary to achieve cross-lingual fine-tuning on low-resource data. 

The texts we use for the following experiments are disjunct from any training data used. Human speech as gold standard is not used, since we are interested in the difference in performance between the systems, not their absolute performance. The close to state-of-the-art performance of the baselines is considered as given, considering their ideal training conditions and use of proven methods. Furthermore, we chose to use German as our benchmark language over an actual low-resource language, since it is much easier to acquire reliable ratings on intelligibility and naturalness for German, than it would be for an actual low-resource language.

\subsubsection{Intelligibility} 
To compare intelligibility between our baseline models and our low-resource models, we use the word error rate (WER) of an automatic speech recognition system (ASR) as a proxy. We synthesize 100 sentences of German radio news texts taken from the DIRNDL corpus \cite{bjorkelund2014extended} with each of our baselines and corresponding low-resource systems. Table \ref{tab:wer} shows WERs that the German IMS-Speech ASR \cite{denisov2019ims} achieves on the synthesized data. For both Tacotron 2 and the FastSpeech 2 based system, the WER of the low-resource model is slightly lower than that of the baseline, thus the low-resource models performed slightly better. 

\begin{table}[th]
    \centering
        \begin{tabular}{l|l l}
            WER & Baseline & Low-Resource \\
            \hline
            Tacotron 2       & 13.1\% & 12.7\% \\
            FastSpeech 2      & 9.9\% & 9.7\% \\
        \end{tabular}
    \caption{WER of the synthesis systems on 100 radio news texts measured using the IMS-Speech ASR.}
    \label{tab:wer}
\end{table}

Looking into the cases where the low-resource system outperformed the baseline, we find code-switched segments, where the texts contain names of Russian cities. Since the pretraining data of the low-resource model includes Russian speech, it seems to have not forgotten entirely about what it has seen in the pretraining phase, which in our interpretation confirms the effectiveness of the LAML against the catastrophic-forgetting problem \cite{french1999catastrophic} of regular pretraining.

\subsubsection{Naturalness}
\begin{figure}[th]
    \centering
    \subfigure[Preference ratings for 102 Tacotron 2 samples.]{\includegraphics[width=.6\linewidth]{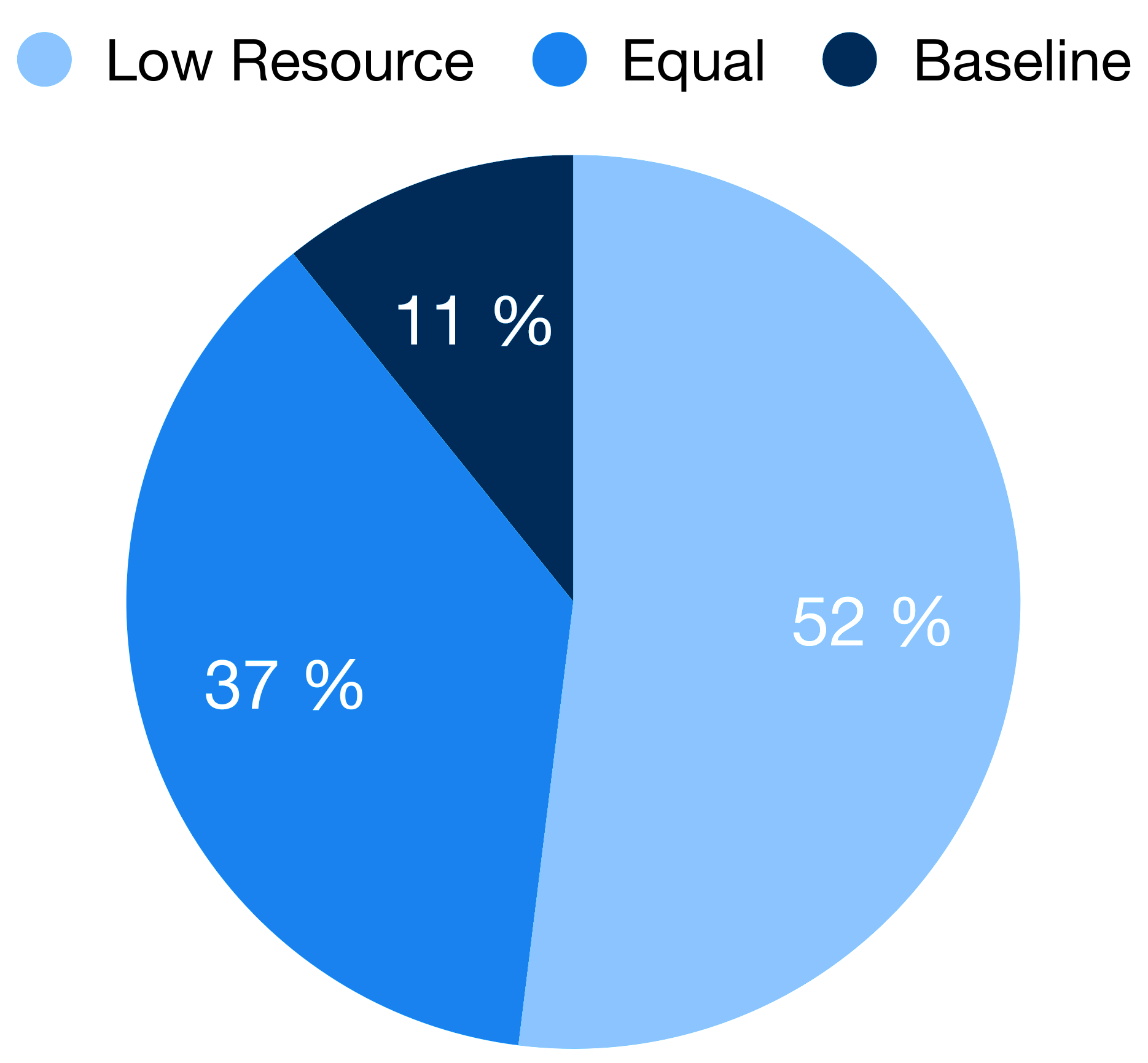}} \\
    \subfigure[Preference ratings for 102 FastSpeech 2 samples.]{\includegraphics[width=.6\linewidth]{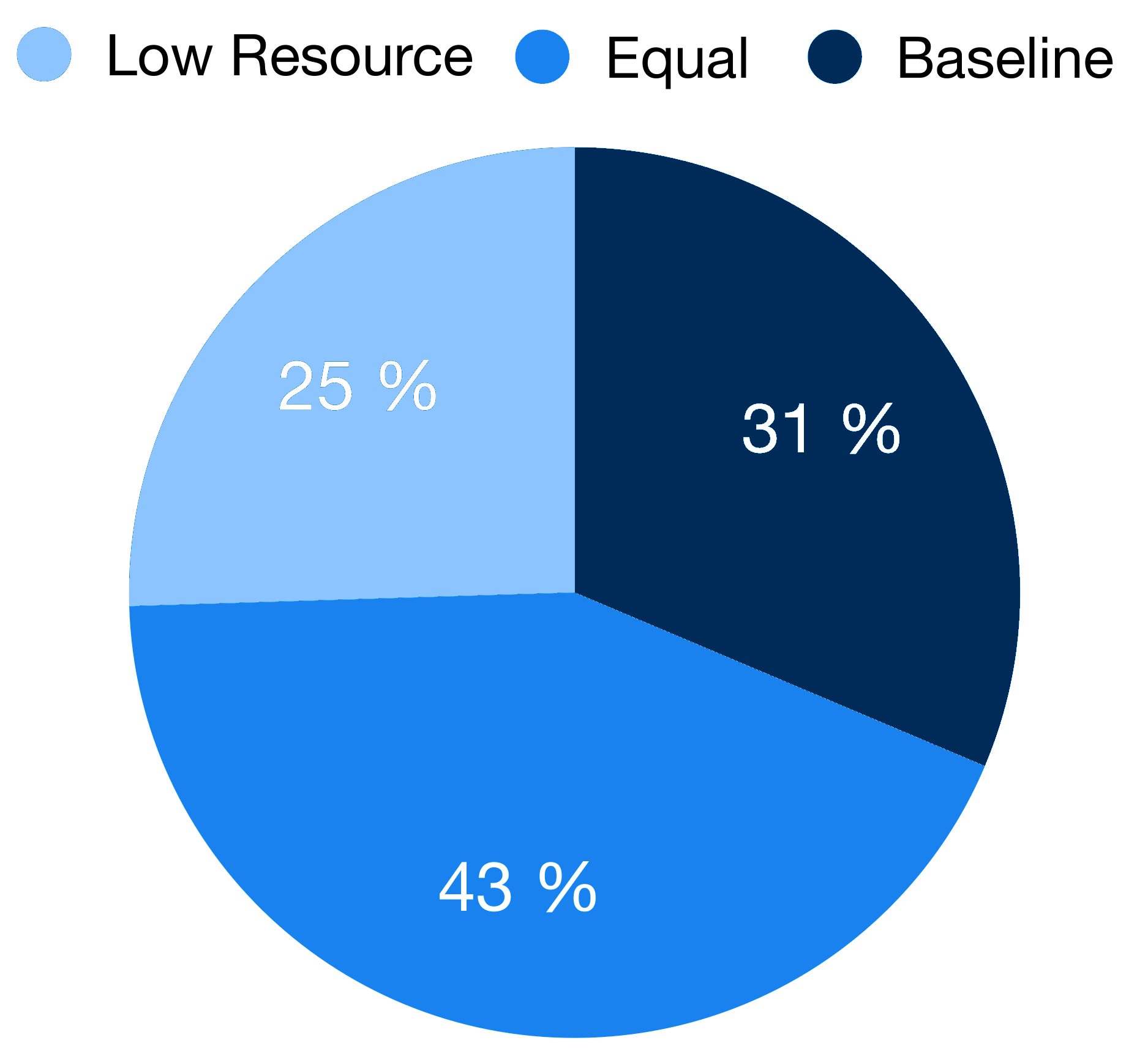}}
    \caption{Results of the preference study comparing a low-resource model to a high-resource baseline.}
    \label{fig:study}
\end{figure}

In order to assess the naturalness of the fine-tuned models, we conduct a preference study with 34 native speakers of German. Each participant is shown 12 phonetically balanced samples produced by the Tacotron 2 and FastSpeech 2 models. For every sentence, there is one sample produced by the baseline and one by the low-resource model. The participants are then asked to indicate their subjective overall preference between the two samples. The results for Tacotron 2 are shown in figure \ref{fig:study} (a). The low-resource system was the preferred system in more than half of the cases, with an equal rating taking up more than another third, showing a clear preference for the low-resource model over the baseline. The results for FastSpeech 2, as seen in figure \ref{fig:study} (b), are a lot more balanced. While the baseline is preferred more often than the low-resource variant, it is not the case in the majority of the ratings. In 56\% of the cases, the model fine-tuned on 30 minutes of data was perceived to be as good or better than the model trained on 29 hours.

\paragraph*{Computational Resources} All models were trained on a single NVIDIA A6000 GPU. Training the Tacotron Baseline took 2 days. Training time of the FastSpeech Baseline was 1 day. Training time of the meta-checkpoint was 4 days, finetuning to a new model from the meta-checkpoint however only takes 2 hours. The HiFi-GAN vocoder used to generate all samples took 4 days to train and was not fine-tuned on the unseen data. We did not perform hyperparameter searches and used the suggested default settings for all methods, which worked sufficiently well, but could surely be improved.  

\section{Further Analysis and Future Work}
\begin{figure}[th]
    \centering
        \includegraphics[width=\linewidth]{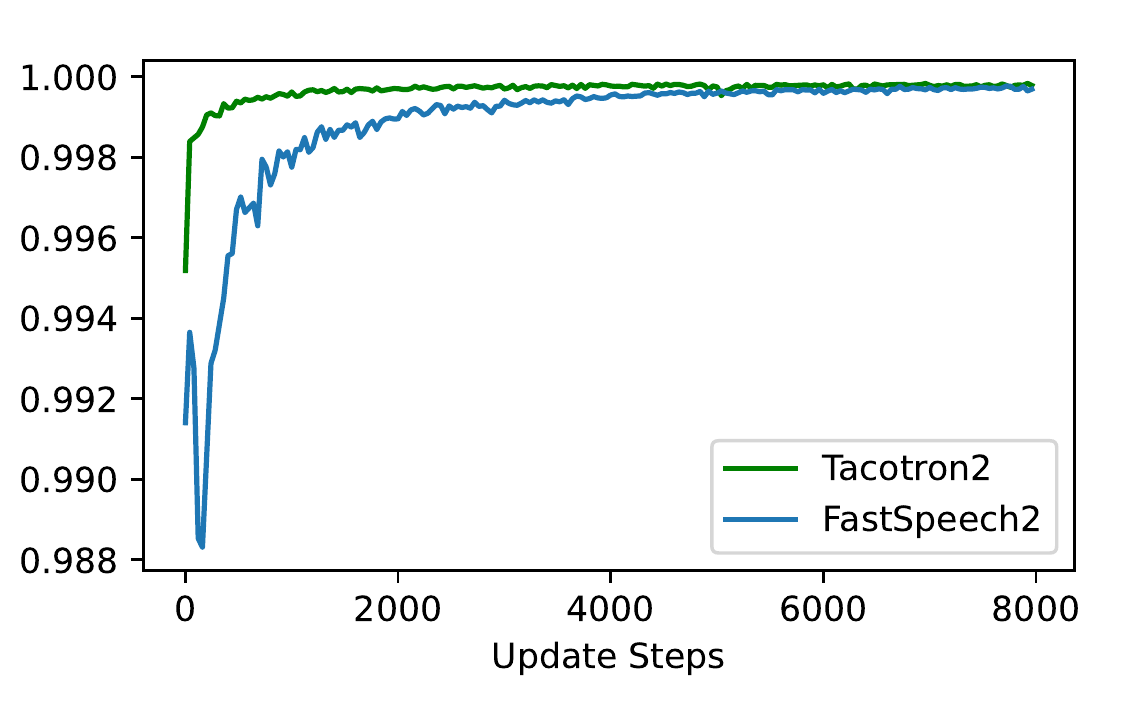}
    \caption{Cosine similarity of speaker embeddings to target speaker over time.}
    \label{fig:spksim}
\end{figure}

\paragraph*{What is the ideal amount of training steps for fine-tuning?}
To investigate the amount of update steps needed to fully adapt to the new speaker with the added difficulty of learning a new language, we show the cosine similarity of a speaker embedding of the fine-tuned model to that of the ground truth throughout the fine-tuning process in figure \ref{fig:spksim}. The speaker embedding is built according to the ECAPA-TDNN architecture \cite{DesplanquesTD20} and provided open source by SpeechBrain \cite{speechbrain}. It is trained on VoxCeleb 1 and 2 \cite{Nagrani17, Nagrani19, Chung18b} which to the best of our knowledge does not overlap with any of the other training and evaluation data we used. We tried to decrease adaptation time further by incorporating said speaker embedding similarity as an additional objective function, similar to \citet{nachmani2018fitting}, we did however see only marginal improvements in the amount of steps needed at the expense of greatly increased training time.

\begin{figure*}[ht!]
    \centering
    \subfigure[66 dimensional Featurespace (24 PanPhon and 42 Papercup features)]{\frame{\includegraphics[width=.3\linewidth]{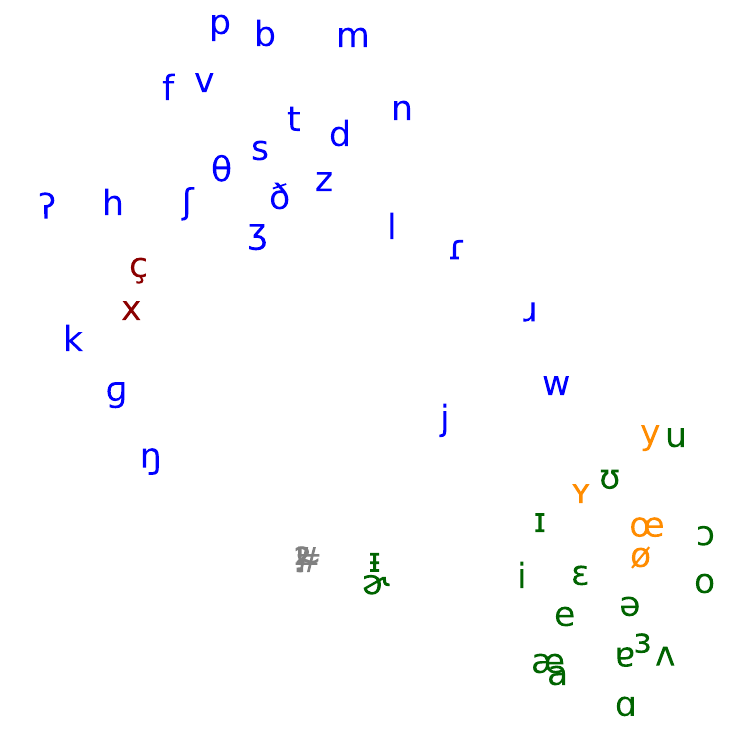}}} \hfill
    \subfigure[512 dimensional embeddingspace learned during Tacotron 2 training]{\frame{\includegraphics[width=.3\linewidth]{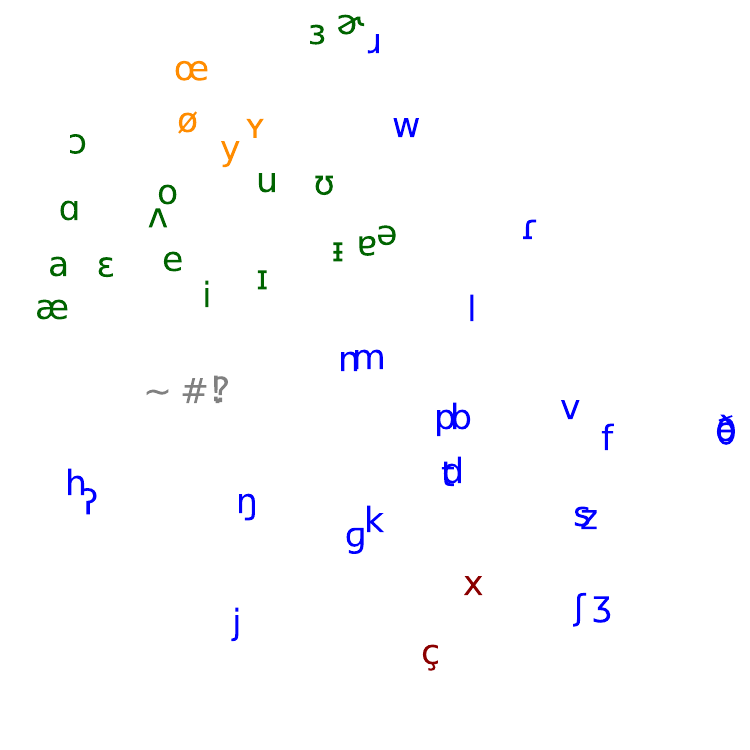}}} \hfill
    \subfigure[384 dimensional embeddingspace learned during FastSpeech 2 training]{\frame{\includegraphics[width=.3\linewidth]{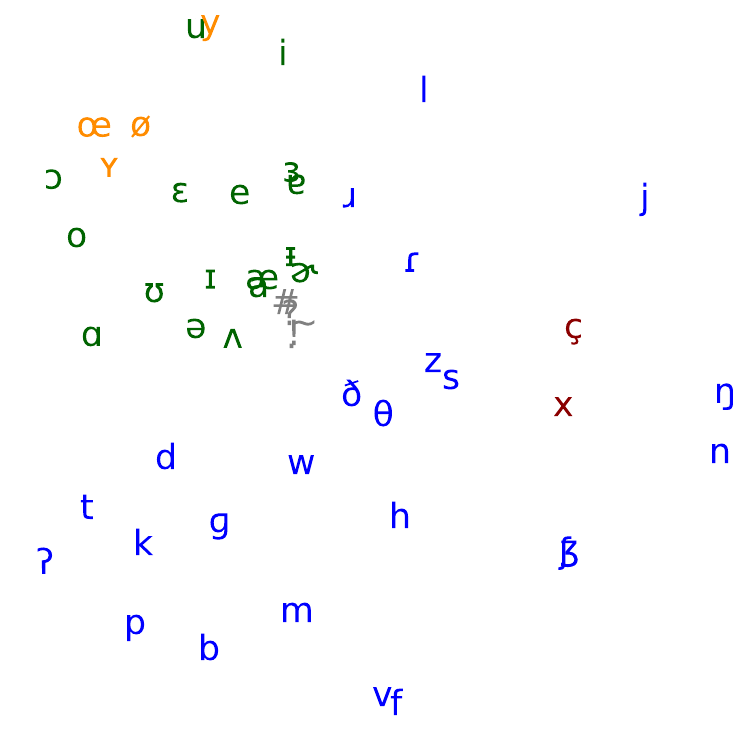}}}
    \caption{t-SNE visualizations of phoneme representations, illustrating zero-shot capabilities. Special characters are grey, consonants are blue, vowels are green, unseen consonants are red and unseen vowels are orange.}
    \label{fig:zero}
\end{figure*}

\paragraph*{Can this setup handle zero-shot phonemes?} We show the model's zero shot capabilities in figure \ref{fig:zero}. We removed Dutch and Finnish from the training data of the meta-checkpoint and trained another version of it, to be able to see how it handles all of the now completely unseen phonemes specific to German. While their correct position in plot (a) can be considered given, since it shows the articulatory featurespace, their meaningful positions in plot (b) and (c)  show that the meta-checkpoint does not just collapse the vector of the unseen phoneme to the one it is most similar to, but actually generalizes. While their pronunciation when produced does not match the correct pronunciation perfectly, it can be understood in the context of a longer sequence. This is congruent with the results of \citet{Staib_2020}. During the adaptation phase, the pronunciation of the unseen phonemes rapidly matches the correct pronunciation after less than 100 steps.

\paragraph*{Does this setup learn the difference between language and speaker?} When analyzing the fine-tuned meta-checkpoint, we observed that it seems to link the language of the input to the voice of the speaker. For example when synthesizing an unseen Hungarian text using Tacotron 2, the voice of the synthesis resembles that of the Hungarian female speaker, even though the model has been fine-tuned on the male German speaker and there are no additional conditioning signals. We hypothesize that the LAML procedure induces certain subsets of parameters in the model to be speaker dependent and the encoder of the model priming those parameters purely based on the phoneme sequence. This leads us to believe, that the fine-tuning of all parameters in the model may neither be necessary, nor even the best way of adapting to new data. This also fits the observations of the speaker embedding over time, since the Tacotron model adapts to the new speaker very rapidly. Further investigations into the interactions between parameter groups could allow cutting down the amount of parameters that need to be trained significantly, further reducing the need for training data. 

\paragraph*{How can we bring down FastSpeech 2's data need further?} A similar observation regarding language and speaker can be made with FastSpeech 2, however as could be seen from the experiment on naturalness and the training time, the FastSpeech 2 model can benefit more from additional data and training time. This may come down to its nearly twice as high parameter count. So a more effective fine-tuning strategy, that considers some parameters as constants, could benefit the fine-tuning capabilities of the FastSpeech 2 model greatly.

\paragraph*{Does this work across language families?} One limitation to our findings is that we investigated only the transfer of languages that share similar phoneme inventories. It is possible that fine-tuning to a language that uses e.g. the lexical tone rather than pitch accents or word accents would require pretraining in more closely related high-resource languages, such as Chinese. However, as \citet{vu2013multilingual} find in their analysis of multilingual ASR, the fast adaptation of an acoustic model trained on multiple languages to unseen languages works well, even across different language families. We thus believe that the technique and analysis presented in this paper also holds across language families and types.

\section{Conclusion}
In this paper, we show an approach for training a model in a language for which only 30 minutes of data are available by making use of articulatory features and language agnostic meta learning. The main takeaways from our work are as follows:

\paragraph*{Articulatory Features for TTS} Using articulatory features as the input representation to a TTS system enables the use of multilingual data without the need for increased architectural complexity, such as language specific projection spaces. It is furthermore beneficial to use even in single-language scenarios, since the knowledge sharing between phonemes makes the TTS system converge much earlier to an usable state during training. 

\paragraph*{MAML on TTS} Applying MAML to TTS does not work well. If we however remove the inner loop, we are able to pretrain a low-resource capable checkpoint for TTS. This modification not only makes it work, it also simplifies the formulation.

\paragraph*{Zero-shot capabilities} The use of articulatory features enables zero-shot inference on unseen phonemes. This is further enhanced by the LAML training procedure. The implications of this are particularly interesting for codeswitching, as \citet{Staib_2020, dan} have pointed out previously. Using these two techniques in conjunction could be used to reduce the problem of codeswitching to a problem of token-wise language identification.

\section*{Acknowledgements}
We would like to thank the anonymous reviewers for their insightful feedback and suggestions.
This work was funded by the Carl Zeiss Foundation.
\bibliography{main}
\bibliographystyle{acl_natbib}

\end{document}